\documentclass[conference]{IEEEtran}
\IEEEoverridecommandlockouts
\usepackage{cite}
\usepackage{amsmath,amssymb,amsfonts}
\usepackage{algorithmic}
\usepackage{graphicx}
\usepackage{textcomp}
\usepackage{xcolor}
\usepackage{comment}

\def\BibTeX{{\rm B\kern-.05em{\sc i\kern-.025em b}\kern-.08em
    T\kern-.1667em\lower.7ex\hbox{E}\kern-.125emX}}

\usepackage{algorithm}
\newcommand\sr[1]{{\color{black} #1}}
\usepackage[bookmarks=false]{hyperref} 
\newcommand\figref[1]{Figure~\ref{fig:#1}}
\newcommand\secref[1]{Section~\ref{sec:#1}}

\usepackage{tikz,xcolor}

\definecolor{lime}{HTML}{A6CE39}
\DeclareRobustCommand{\orcidicon}{%
	\begin{tikzpicture}
	\draw[lime, fill=lime] (0,0) 
	circle [radius=0.16] 
	node[white] {{\fontfamily{qag}\selectfont \tiny ID}};
	\draw[white, fill=white] (-0.0625,0.095) 
	circle [radius=0.007];
	\end{tikzpicture}
	\hspace{-2mm}
}
\foreach \x in {A, ..., Z}{%
	\expandafter\xdef\csname orcid\x\endcsname{\noexpand\href{https://orcid.org/\csname orcidauthor\x\endcsname}{\noexpand\orcidicon}}
}

\IEEEoverridecommandlockouts
\IEEEpubid{\begin{minipage}[t]{\textwidth}\ \\[10pt]
        \centering\normalsize{978-1-7281-4701-7/20/\$31.00 \copyright 2020 IEEE}
\end{minipage}}

\begin{document}

\title{Deep Reinforcement Learning for Long Term Hydropower Production Scheduling\\
}

\author{\IEEEauthorblockN{Signe Riemer-Sørensen \orcidS{}}
\IEEEauthorblockA{\textit{Mathematics and Cybernetics} \\
\textit{SINTEF Digital}, Oslo, Norway \\
signe.riemer-sorensen@sintef.no}
\and
\IEEEauthorblockN{Gjert H. Rosenlund \orcidG{}}
\IEEEauthorblockA{\textit{Energy systems} \\
\textit{SINTEF Energy Research}, Trondheim, Norway \\
gjert.rosenlund@sintef.no}
}

\maketitle

\begin{abstract}
We explore the use of deep reinforcement learning to provide strategies for long term scheduling of hydropower production. We consider a use-case where the aim is to optimise the yearly revenue given week-by-week inflows to the reservoir and electricity prices. The challenge is to decide between immediate water release at the spot price of electricity and storing the water for later power production at an unknown price, given constraints on the system. We successfully train a soft actor-critic algorithm on a simplified scenario with historical data from the Nordic power market. The presented model is not ready to substitute traditional optimisation tools but demonstrates the complementary potential of reinforcement learning in the data-rich field of hydropower scheduling.
\end{abstract}

\begin{IEEEkeywords}
machine learning,
expert systems,
power generation economics,
hydroelectric power generation
\end{IEEEkeywords}

\section*{Nomenclature}
\begin{description}
    \item[$a_i$] Action, amount of water to convert to electricity as percentage of maximum production capacity.
    \item[$i$] Net inflow, normalised with $r_\mathrm{max}$.
    \item[$f_\mathrm{max}$] Maximum production relative to reservoir capacity.
    \item[$k_\mathrm{price}$] Importance of price in the reward function.
    \item[$q_\mathrm{price}$] Power of price in the reward function.
    \item[$R_i$] Reward for action $a_i$, arbitrary units.
    \item[$r_\mathrm{max}$] Volume of reservoir (M$m^3$).
    \item[s] State, characterised by week number, reservoir level, weekly inflow and weekly price.
    \item[$y_i$] Weekly price normalised to maximum (M$m^{-3}$).
\end{description}

\section{Introduction}
\subsection{Motivation and background}
\begin{figure}\centering
\includegraphics[width=0.49\textwidth]{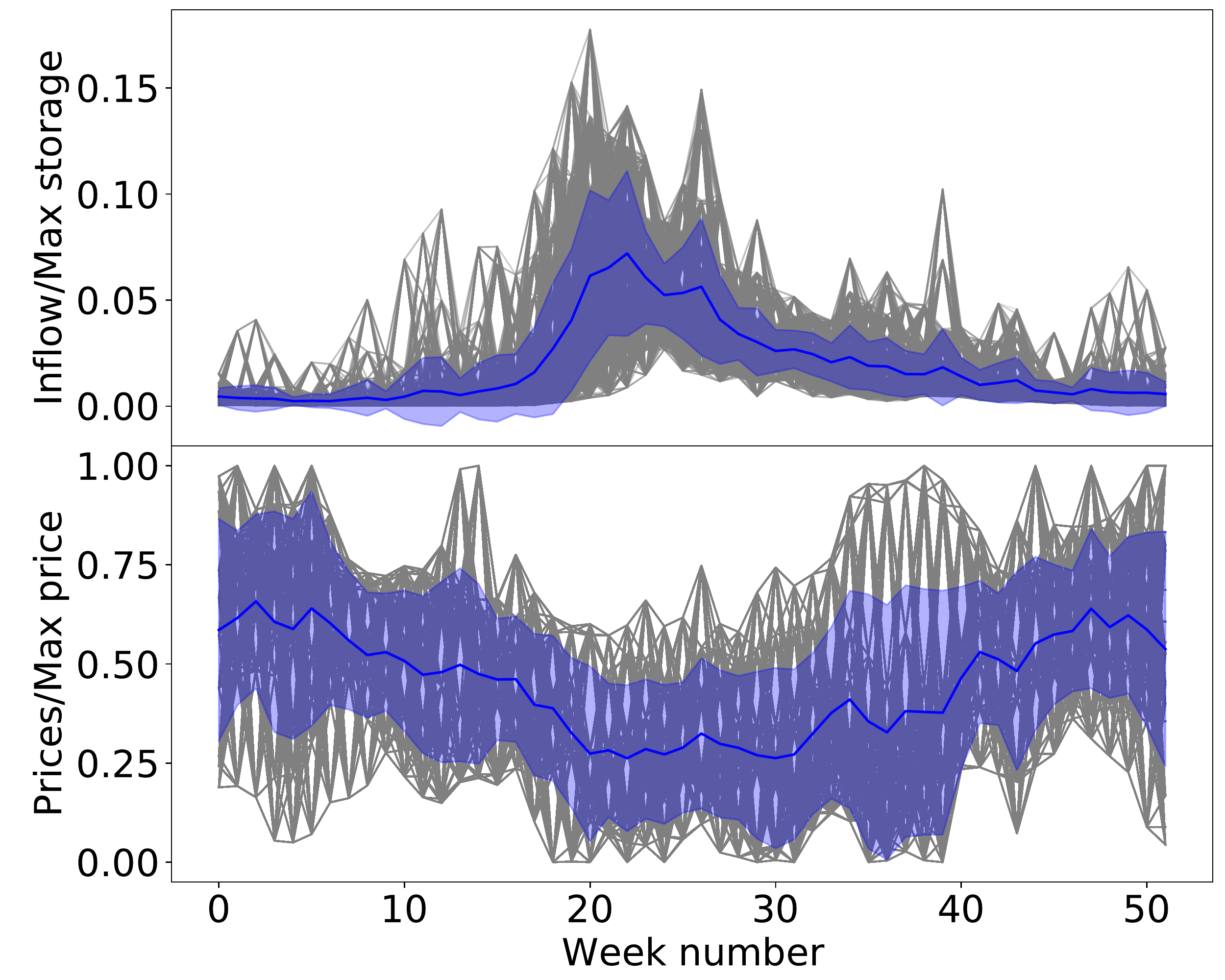}
\caption{7975 samples of historic inflows and prices as a function of week number for four similar water systems in Southern Norway: Tora, Storeskar, Sula, Rinna. The solid blue line indicates the average and the shaded region is the standard deviation. Inflow is highest during spring and early summer, while prices are on average highest during the winter.}
\label{fig:inflows_prices}
\end{figure}

Hydroelectricity is sustainable energy, but due to seasonal variations, the natural inflow to the reservoirs does not follow demand or price as shown in \figref{inflows_prices}. In order to maximise revenue and minimise water spillage, the power generator schedule production based on a combination of deterministic and stochastic models of the inflow and demand \cite{WOLFGANG20091642,Helseth2016}. In traditional methods, the size and complexity of the computations requires human intervention to decompose the problem into smaller parts. By utilising advancements in reinforcement learning \cite{Sutton}, we take a step towards automating the process and solving the problem without human intervention.

In the concept of reinforcement learning a reward scheme is used to train an agent (the algorithm) to a desired behaviour (action) in a range of situations (states) within a given \sr{system (the environment)}. Reinforcement learning is most efficient when the environment and \sr{its response to actions taken by the agent (rewards) are governed by some underlying patterns that are at least partly deterministic. If there are no relations between actions and rewards, there is nothing to learn (e.g. chapter 3.1 in \cite{Sutton}).} Thus, the goal is for the algorithm to generalise over stochastic variations.

In the case of hydropower scheduling, the options for the actions are well defined (release water or save), while the feedback from the environment is a mixture of deterministic and stochastic: Prices are related to production and follow a yearly pattern but with fluctuations, inflows are weather dependent (stochastic) with general seasonal variations. 

Recent years have seen a massive development within the combination of deep learning and reinforcement learning, enabling algorithms to solve complex tasks \cite{Mnih2015, Silver2016, Lillicrap2015}. They are highly flexible and powerful methods that potentially can be adapted to many different purposes but given their complexity, very few applications have been demonstrated outside simulated or strictly rule based systems \cite{rlblogpost}.

\subsection{Contributions and organisation}
\sr{Our contribution is to modify and train a state-of-the art soft actor-critic algorithm on a toy scheduling scenario and two realistic scenarios, to demonstrate the potential of reinforcement learning in hydropower scheduling. As will be explained in \secref{sac}, the soft actor-critic algorithm} is particularly well suited to systems with a high degree of stochastic fluctuations. The agent is trained in a ``safe'' environment made up from stochastic variations of historical data (\secref{usecases}), from which it is able to successfully learn a meaningful policy for water release. The algorithm is not (yet) a replacement for traditional optimisation and planning methods, but rather we discuss its potential as a complementary method in \secref{discussion}.

\section{Soft actor-critic reinforcement learning} \label{sec:sac}
Soft actor-critic reinforcement learning is based on Q-learning, where all states and actions are assigned a value \cite{Watkins1989, Watkins1992}. The aim is to learn an approximation for the value of \sr{all} possible action\sr{s} in each state, and for each step choose the action \sr{with} the potential to provide the largest total reward. In its pure form, Q-learning is limited to discrete states and actions, \sr{it} is not very flexible and scalable, and not very robust to stochastic fluctuations and time-dependent evolution \cite{Sutton}.

In actor-critic algorithms, the Q-learning concept is expanded and the policy and value functions are separated. \sr{The value function is used in the learning process (the critic), but the actor selects the action without consulting the value function}.
In the soft actor-critic algorithm, the lifetime reward optimisation is combined with an entropy maximisation \cite{Haarnoja2018} leading to substantial improvement in terms of learning speed, final performance, sample efficiency, stability, and scalability, in particular on complex tasks or on stochastic systems\cite{Haarnoja2018b}.

\section{Implementation} \label{sec:implementation}

\begin{figure*}\centering 
\includegraphics[width=0.80\textwidth]{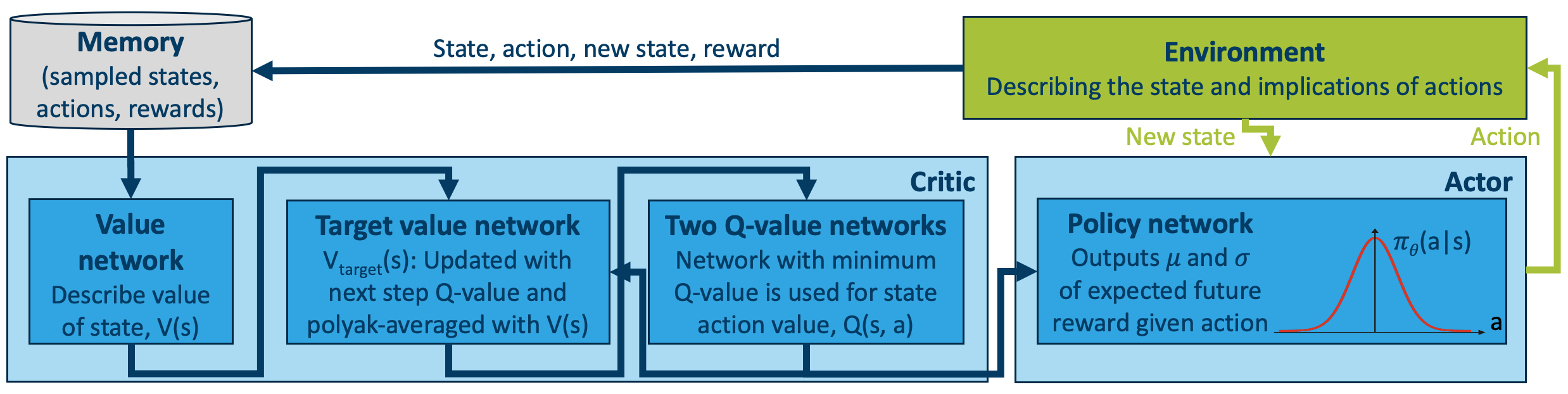}
\caption{The components of the soft actor-critic algorithm and their interplay.}
\label{fig:sac}
\end{figure*}
The algorithm is implemented with a policy network (the action to take for a given state), a value network (describing how advantageous each state is), a target value network, and two soft Q-networks \cite{Haarnoja2018} \sr{as illustrated in \figref{sac}}. 

\sr{The target value network is introduced because the target for training the Q-network depends on the value network, and the target for the value network depends on the Q-network. Consequently, this loop makes the training unstable. The solution is to use an additional network while training the Q-network, which is close to the value network but with a short time delay. Thus, the target value network becomes a memory of how the value network was a moment ago. Instead of copying the value network directly, we perform a Polyak averaging between the target value network and the value network to obtain a kind of moving average over the gained experience \cite{Polyak1992}.}

The theory only calls for one Q-function, but in practice, there is a tendency for the algorithm to overestimate the Q-values. \sr{This is mitigated by training two Q-networks and always using the minium of} the two values when updating \sr{the} policy and value function networks. \sr{Neither is it} strictly necessary to have separate approximators for the value and Q-functions, since they are related through the policy, but in practice it help\sr{s} with convergence. \cite{Haarnoja2018b} \sr{discusses how the value function can be disposed of and introduces automatic detection of} the weight of the entropy term (called the temperature). 

We \sr{apply} memory replay where the learning experiences are stored and reused when training the neural networks, in order to obtain a better balance between rare and common events.

We follow the pytorch implementation of \cite{Kumar2018, NEURIPS2019_9015} but with significant modifications related to the activation functions and determination of accessible states.

\subsection{The environment} \label{sec:environment}
While the algorithm itself is generic, the environment and reward functions are specific to the problem we want to solve. \sr{In its simplest form, the purpose of hydro power scheduling is to maximise the income, given scenarios on inflow. This becomes a balancing act between storing the water for winter (high prices), while not keeping too much, leading to spillage during spring (high inflow, snow melting).}

The reservoir \sr{holds a maximum available volume for production of $r_\mathrm{max}$, and $i$ accounts for the net inflow}. If the capacity of the reservoir is exceeded, the water will overflow as spillage. All water volume quantities are given in M$m^3$, and \sr{in the algorithm they are normalised to maximum available water volume so they become unit-less.} 

Each state is defined by week number, storage, price for the week, inflow for the week, and the number of weeks it would take to empty the reservoir if running on full capacity. The latter is used to clamp the policy function to \sr{feasible} actions only. The initial storage in the reservoir is randomly sampled from a specified distribution.

\sr{We consider the simplest possible viable version of the problem}. The method scales well and \sr{further} constraints can be implemented with relatively low effort \sr{(see \secref{discussion}).}

The training data (described in \secref{data}) are also loaded into the environment. The inflow is normalised to maximum \sr{reservoir} volume, and the prices are normalised to the maximum price. In addition, \sr{the water remaining in the reservoir at the end of the year can be given a value e.g. the value from the last week, or the average across all scenarios}. 

\subsection{The action space}
The action space is defined as the amount of water to be released \sr{and converted to electricity} at the spot price. Normalising with the maximum production capacity, this becomes a continuous variable \sr{with values between 0 and 1}. Since the volume never can be a negative number, we have implemented a sigmoid activation function and scaling in the policy network. 

\subsection{The step function and reward}
\sr{The step function determines the response by the environment of a given action performed in a given state. This includes computing the reward and the transition to the next state (update reservoir level and informing of new inflow and price values).}
In order to make the algorithm as transparent as possible, we have maintained a fairly simple reward determination: \sr{First the suggested production and corresponding end storage is computed. Then we check the feasibility of the action.} If \sr{the} end storage is larger than \sr{the} capacity, the reservoir is flooding and the end storage is adjusted to maximum capacity. 

The reward $R_i$ for action $a_i$ is given by:
\begin{equation}\label{eqn:reward}
R_i = a_i f_\mathrm{max} r_\mathrm{max} (y_i k_\mathrm{price})^{q_\mathrm{price}},
\end{equation}
where $f_\mathrm{max}$ \sr{(unit-less)} is the maximum production capacity \sr{(maximum volume of water that can be converted to electricity)} relative to the total reservoir capacity. $r_\mathrm{max}$ is the reservoir capacity \sr{(M$m^{3}$)}. $y_i$ is the weekly price normalised to maximum price (M$m^{-3}$) and the importance of the price is controlled with $k_\mathrm{price}$ and $q_\mathrm{price}$. $k_\mathrm{price}$ can also be used to relate the produced volume to actual prices \sr{and account for a non-linear production function. The reward is a unit-less number unless otherwise adjusted with $k_\mathrm{price}$.}

\subsection{The networks}
The value networks (both main and target), and the soft Q-networks are standard fully connected \sr{neural} networks each with three hidden layers with relu activation functions \cite{Hahnloser2000} and an output layer with a linear activation function. The number of neurons in all hidden layers is a hyperparameter to be set by the user.

The policy network has two hidden layers with relu activation functions. It has two outputs, the mean and the logarithm of the standard deviation (clamped to be in a sane region). These are used for a re-parametrisation that ensures that the sampling from the policy is differential and the errors can be properly back propagated, leading to faster convergence. 
The action to be taken from a given state, $s$, is obtained from the policy function by sampling noise, from a standard normal distribution, multiply it with the standard deviation, add it to the mean, and then activate it with a sigmoid activation function to ensure an action between 0 and 1 \cite{Haarnoja2018}.

The full update process then becomes:

\begin{enumerate}
	\item Initialisation of the networks.
	\item Initialisation of the environment.
	\item Training loop:
	\begin{itemize}
		\item At the beginning of each episode (full year), reset the environment and sample the initial amount of water in storage.
		\item For each week, randomly sample the price and inflow, and obtain the action from the policy (or randomly in the early exploration phase).
		\item Determine the reward and the next state (step function in environment).
		\item Save experience to memory.
		\item Update networks in batches from memory:
		\begin{itemize}
			\item Predict values of Q-functions, value and policy network (for all states in the batch and their suggested actions).
			\item Evaluate the policy network to get the next states.
			\item Predict the target value network.
			\item Compute loss of Q-functions and value network and do one back propagation (update weights).
			\item Adjust the target value function with next step Q-value. 
			\item Compute the loss of the target value function and do one back propagation.
			\item Compute loss of policy network and do one back propagation.
			\item Update target value network by Polyak averaging with the main value network.
		\end{itemize}
		\item Save the model.
	\end{itemize}
\end{enumerate}

\subsection{Hyperparameters}  \label{sec:hyperparameters}
The network structures (number of layers) are hardcoded and not considered hyperparameters here. However, the following parameters must be chosen (the values in parentheses indicate the values applied in \secref{usecases}):
\begin{itemize} 
\item Number of neurons in the hidden layers (100).
\item Absolute values for the outer edges of the range of uniform initialisation weights ($3\times 10^{-3} \in [0, \infty]$).
\item Loss function for the value and soft Q networks (mean squared error).
\item Optimisation functions for all networks (RMSprop).
\item Learning rates of the value network, the soft Q networks and the policy networks ($5\times 10^{-4}$, $5\times 10^{-4}$, $1\times 10^{-4}$).
\item Clamp values for the logarithmic standard deviation on the policy network. The results are insensitive to the exact values (-20, 2).
\item Minimal logarithmic probability. The results are insensitive to the exact value ($3\times 10^{-6}$).
\item Discount factor, $\gamma$ \sr{($0.99 \in [0, 1]$)}.
\item Soft $\tau$, determines the importance of the main value network when updating the target value network \sr{(0.0006)}. 
\item Number of exploration steps before the training begins (10000 in the artificial case and 50000 in the historic).
\item Total number of weeks to train (300000, model may converge before).
\item Number of experiences to save in the replay memory (all).
\item Batch size to use from replay memory (100). The number of randomly selected experiences to use for every update of the networks.
\end{itemize}

We use RMSprop \cite{Tieleman2012} as optimiser because its lower dependence on momentum than e.g. adam \cite{Kingma2014}, makes it more adaptable and well suited to handling the non-stationary data distribution from the changing environment. In addition, the policy network has a smaller learning rate than the value functions, in order to collect experience at a faster rate than the policy adapt to the experience. \sr{While slowing down training, this prevents excessive exploration of local minima.} 

The hyperparameters were decided based on default values from \cite{Haarnoja2018, Haarnoja2018b} and manual tuning. Unfortunately, there are no efficient procedures for obtaining the optimal set of hyperparameters in reinforcement learning. However, several of the parameters will mainly influence convergence rate, so if trained for a large enough number of episodes, the final model performance will be similar.

The environment also contains some \sr{system definitions and} hyperparameters:
\begin{itemize}
\item Maximum capacity of the reservoir\sr{, volume (1000)}.
\item Maximum production, \sr{volume per week (30 and 100)}. 
\item Scaling factor for the price, $k_\mathrm{price}$ (1). Numeric factor that allows for tuning the sizes of the rewards and e.g. relate them to actual prices.
\item Price power, $q_\mathrm{price}$ (1). The importance of the price in the reward function. \sr{Can also be used to account for non-linear conversion.}
\item Initial storage (randomly sampled between 0.4 and 0.6 of full storage capacity).
\end{itemize}

\section{Use cases} \label{sec:usecases}
\sr{Our main purpose is to validate reinforcement learning as a method for solving the problem of seasonal hydropower scheduling. To make the implementation and tuning of hyperparameters transparent, we have designed use cases without the most complicating elements; the production of electricity is assumed to be linear to the amount of water released (but the reward function allows for a power law relationship), the station is not part of a cascaded or otherwise restricted system, and there are no pumps or hatches to operate. Consequently, they serve as a minimal viable demonstration of the algorithm.} 

\sr{A common measure of flexibility in a hydropower plant is the usage time defined as how long the plant would operate at maximum capacity to convert a full reservoir to electricity. The use cases are designed to reflect common usage times for Norwegian power stations, while being sufficiently diverse for the algorithm to learn different strategies.} 

\sr{The first power station} has a relation between maximum production capacity and total \sr{reservoir} volume of 30/1000 corresponding to a usage time of 5600 hours. The average yearly inflow is \sr{taken to be} the same as the reservoir capacity. \sr{The second power station is used to demonstrate the changes when both the inflow and production capacity are higher with a ratio of 100/1000 corresponding to 1680 hours of usage time, and yearly inflow of 4000. For both reservoirs}, the initial storage is randomly sampled between 40\% and 60\% of the full storage capacity.

\subsection{Training data} \label{sec:data}
We train and apply the agent on two \sr{sets of historic data}. Firstly, we train on artificial scenarios, where there is a clear structure in the price and inflow, albeit with \sr{some} fluctuations. In the second case, we use historic price data from the Nordic region (NO3) for the period\footnote{provided by the NordPool Group \sr{\url{nordpoolgroup.com}}} 2008 to 2019 and inflow provided by The Norwegian Water Resources and Energy Directorate for the period\footnote{\url{nve.no/hydrologi/hydrologiske-data/historiske-data/historiske-vannforingsdata-til-produksjonsplanlegging}} 1958 to 2018. \sr{In order to expand the inflow data without introducing additional noise, we combine inflows from four reservoirs in the NO3 region with similar weather patterns: Tora, Storeskar, Sula and Rinna (shown in \figref{inflows_prices})}. The weekly prices and inflows are linearly scaled to the yearly minimum and maximum values. We construct each scenario by randomly sampling 52 pairs of price and inflow values from the corresponding samples for the particular weeks. \figref{results} shows examples of the scenarios in the upper two panels. In the artificial scenarios, the water remaining in the reservoir at the end of the year is valued at its price for the last week \sr{and in the historic scenarios it fetches maximum price}. \sr{The agent is only rewarded for remaining water at the end of the year if the reservoir filling is between 0.4 and 0.6. This is to encourage some storage for the next year. In an operational setting, the best treatment of the end-value condition has to be further investigated.}

\subsection{Results}
\begin{figure}\centering 
\includegraphics[width=0.49\textwidth]{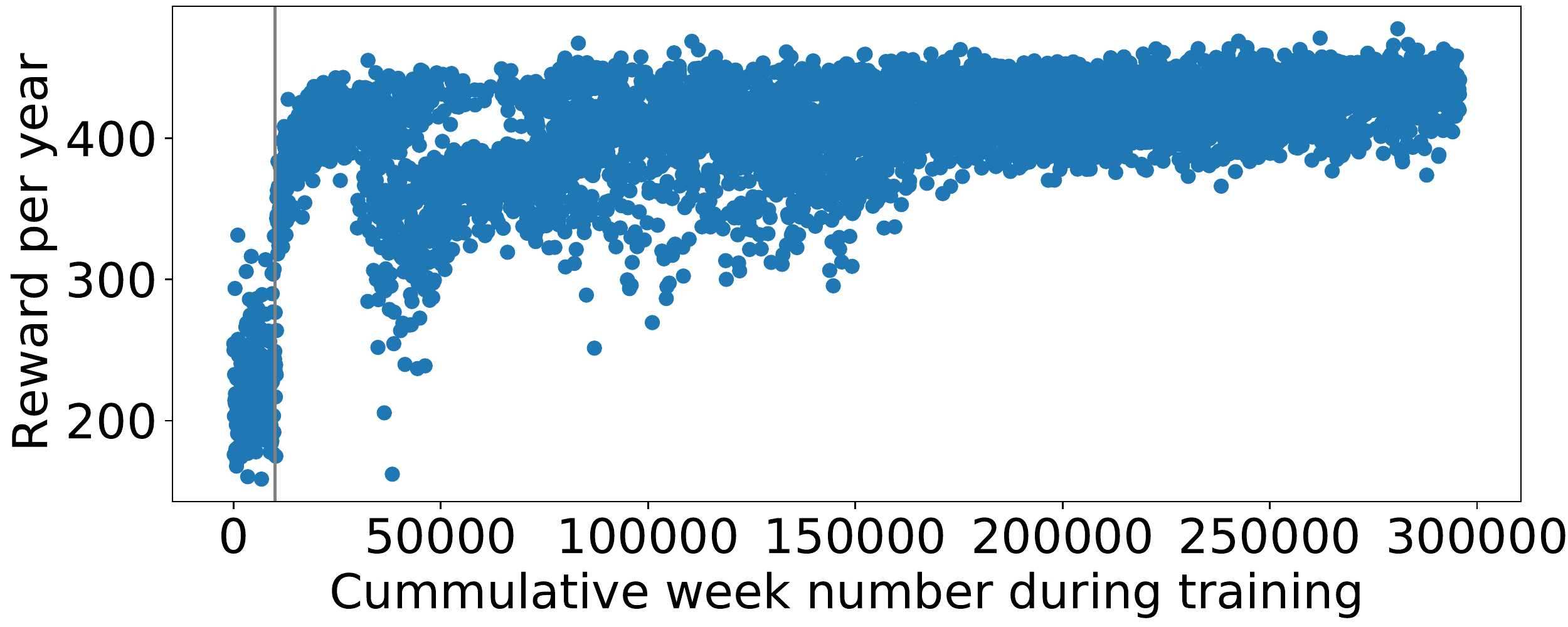}
\caption{Example of total reward as a function of training episode. The vertical line indicates the transition from random exploration to policy driven actions.}
\label{fig:rewards}
\end{figure}

Training the algorithm on the use cases described in \secref{usecases} for \sr{300000 episodes takes a day} on a 3.1\,GHz processor with 16\,Gb RAM available. \sr{This training only needs to be done once for each reservoir/system. Applying the trained model requires less than a minute to provide a plan for 52 weeks. After deployment, the model can be updated continuously with experience gathered from new data (timescale of minutes). For more complex systems (e.g. cascading reservoirs), the initial training is expected to take longer.}
\figref{rewards} shows an example of the total rewards as a function of training episode. The transition from random exploration to policy driven actions \sr{leads to an overall increase in rewards but with continued exploration.}

\begin{figure*}[ht]\centering 
\includegraphics[width=0.49\textwidth]{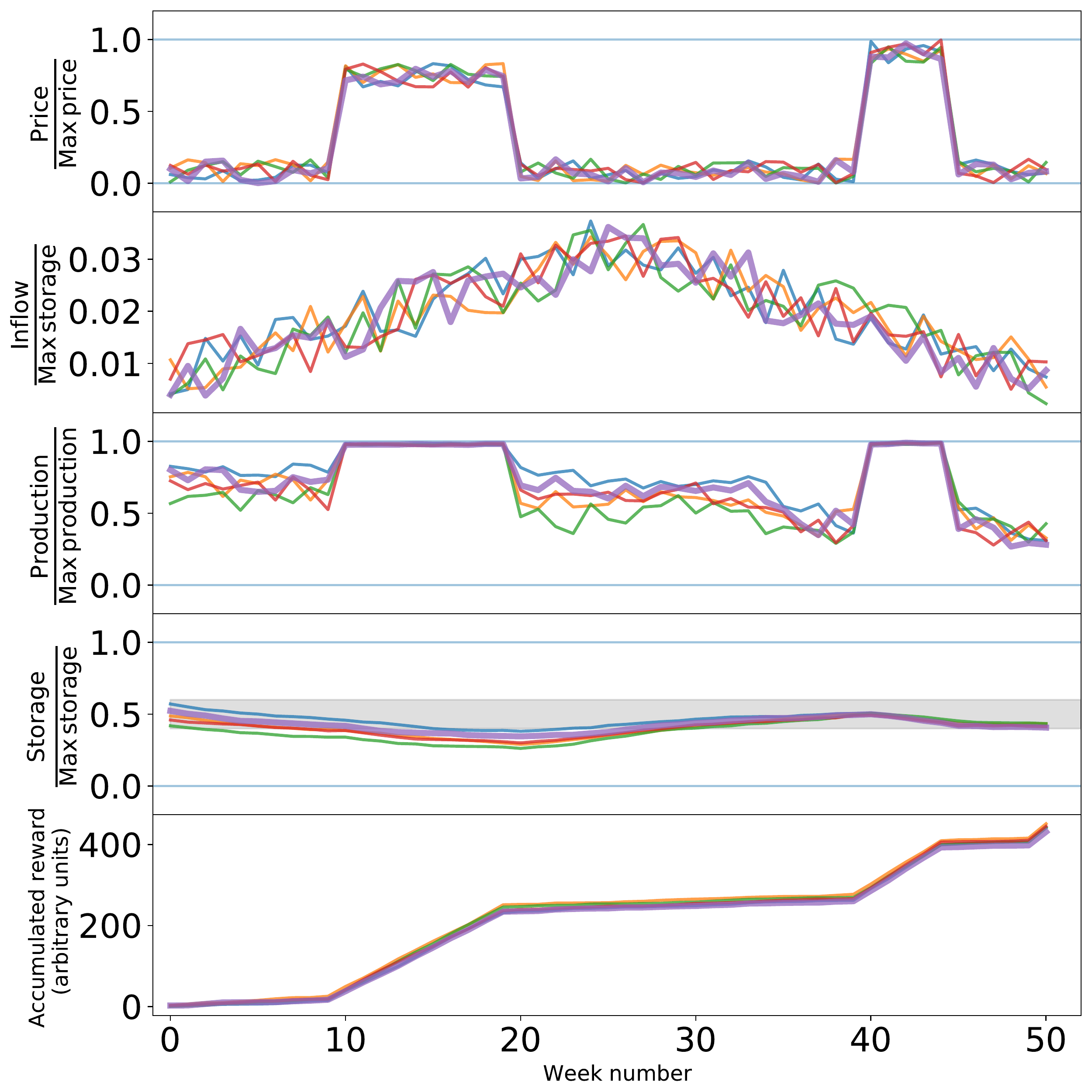} 
\includegraphics[width=0.49\textwidth]{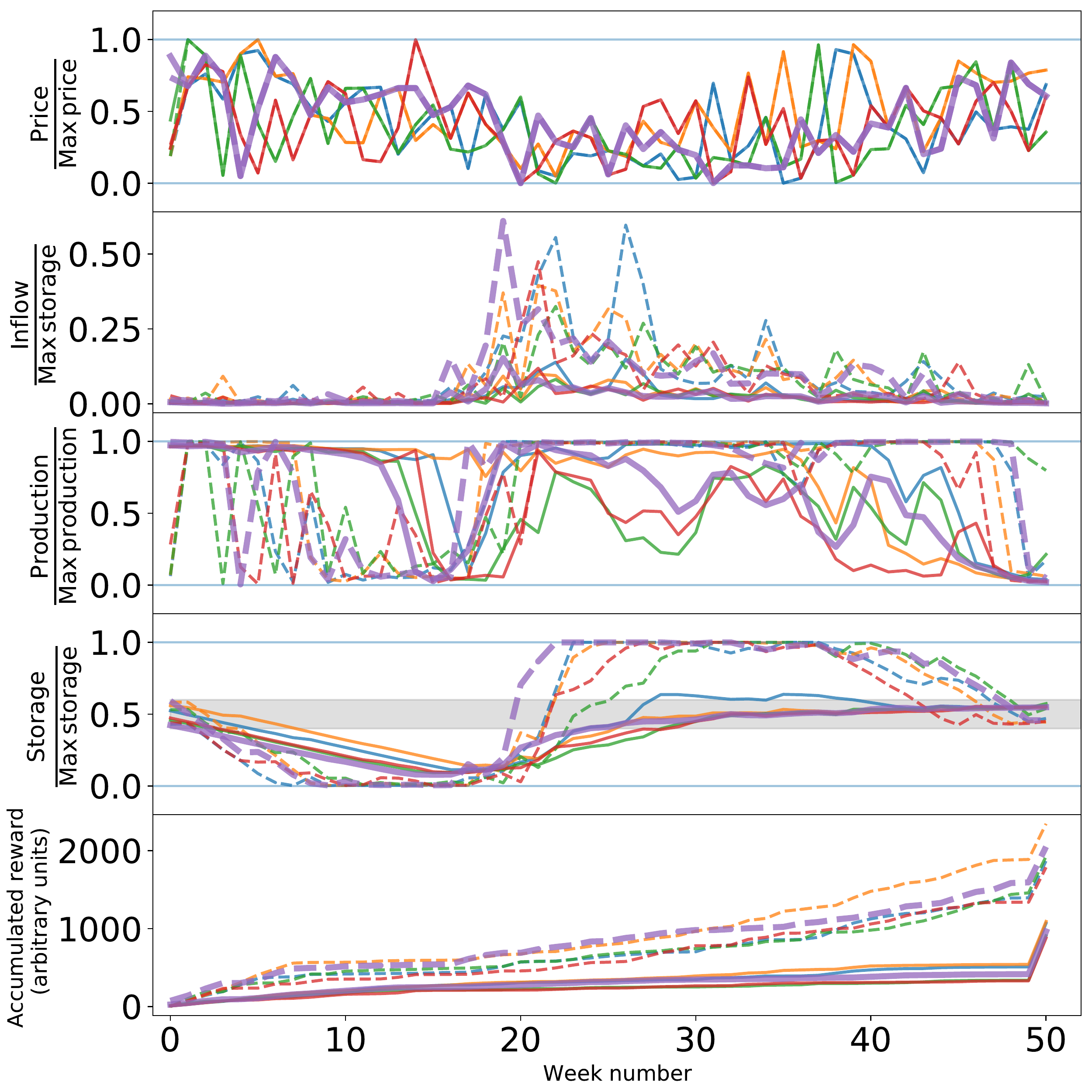}
\caption{Examples of price, inflows, actions, storage and accumulated reward per week for five different scenarios randomly sampled from the training data. {\it Left:} An artificial price scenario with \sr{instantaneous price changes and clearly defined seasonal variation of inflow}. {\it Right:} Prices and inflows based on historic Nordic data \sr{with some seasonal trends but also a large variation}. The solid lines represent a production/storage ratio of 30/1000 and yearly inflow of 1000, and the dashed lines represents 100/1000 and yearly inflow of 4000. \sr{The high inflow scenario has more water available and consequently higher rewards.}}
\label{fig:results}
\end{figure*}

\sr{\figref{results} shows the} resulting actions for five example scenarios for each \sr{fictive reservoir}. In \sr{all cases, the agent was trained for ~3000000 episodes corresponding to ~5700 years of training data. As seen in \figref{rewards}, the model converged earlier and the training could have been stopped}. The agent has successfully learned a meaningful policy for water release given the limited state information.

When the price and inflow data are dominated by fluctuations, the resulting actions will also have a large range of fluctuations. This is partly due to the lack of generalisability for the value function and partly due to the entropy optimisation, where the soft actor-critic agent will automatically optimise the entropy and increase the random exploration for regions of the parameter space with large fluctuations. This behaviour \sr{is} part of the reason why soft actor-critic is ideal for stochastic dominated environments, because it protects against the actor learning to always perform the same action and fail miserably when conditions change. In \figref{results} we show the actions from the trained agent without the additional stochasticity, in order to see how the policy exploits what it has learned. \sr{This approach is known to lead to better performance than including the stochasticity at runtime \cite{spinningup}.}

In the artificial case, the agent has learned to release most of the water when there is a high revenue albeit with some variation to protect against stochastic variation.

In the historic case, the agent releases water in the beginning of the year when the price is high, but less water in second half of the year in order to obtain the reward for having water in storage at the end of the year. \sr{For the system with production capacity of 30/1000, the} storage is not fully exploited as there is no incitement in the reward function or environment for storing more than 60\% of the capacity. \sr{For the system with higher inflow and larger production capacity}, the additional capacity becomes a buffer to hold the inflow.

\section{Discussion and outlook} \label{sec:discussion}
We have demonstrated that the agent is able to learn a policy that is sensible for human interpretation. However, this \sr{reflects a minimal viable case} and at the current level, the policy is not ready for real-life deployment. The resulting production plan is somewhat different from the suggestions of traditional stochastic optimisation tools, indicating that the reinforcement learning may be able to exploit the data differently than classical models. Consequently, we do not advocate that the classical models can be replaced by reinforcement learning, but rather that they can be complementary.

Reinforcement learning has previously been attempted used in hydropower scheduling \cite{Cote2015, Castelletti2009} and in combining multiple connected reservoirs \cite{Lee2007}. \sr{Relative to \cite{Lee2007, Castelletti2009} which both apply Q-learning and consequently discrete state/action spaces, soft actor-critic algorithm allows for continuous state/action spaces and is more stable towards stochastic variation. Another major difference is} the step towards higher degree of realism with the implementation of price/market in the environment. The framework is flexible and the algorithm is easily applied to different time scales or extended with additional constraints e.g. cascading reservoirs, minimum production requirements\sr{, non-linear production functions, }or ramp-up costs. In addition, the neural network structure can be changed to take temporal aspects into account (recurrent neural networks) or structured information in the form of graphs.

\sr{For cascading reservoirs or systems with external water flow restrictions, the constraints must be implemented in the environment. This can either be as hard constraints on allowed actions or via combination into a single reward function (e.g. as in \cite{Lee2007}). For cascading systems, the agent can be given control of the entire system, in which case the action space must be expanded accordingly to include production and active spillage in all units. Due to the increased size of the action space, it is expected that the model training will take longer and the training data must naturally reflect the topology.}

The agent can be further informed by combining the training data with weather forecasts or traditional models for e.g. forecasting inflow and demand. In addition, \sr{one can take advantage of transfer learning \cite{Pan2010}, where models can be pre-trained on data from one region and then transferred to similar regions reducing the training time at each location.}

We see a huge application potential for reinforcement learning in situations where the current solution to handle system complexity is aggregation and dis-aggregation of models. When the entire system becomes too complex and computationally heavy for a single model, the individual reservoirs are aggregated in the model and their joint production scheduled. Reinforcement learning could then be applied for the dis-aggregation of the model and the scheduling for the individual reservoirs given constraints from the aggregated model. It is straight forward to couple reservoirs and add an overall production criteria in the framework.

\section{Conclusion}
We have explored reinforcement learning to provide strategies for long term scheduling for hydropower production. A soft actor-critic agent is able to learn a meaningful policy on both artificial and real-world training scenarios. While the method is not yet perfect, \sr{its flexibility and ability to generalise even complex scenarios gives it a} potential to complement traditional optimisation methods for hydropower scheduling.

\section*{Acknowledgment}
The authors thanks Eivind Bøhn for enlightening discussions, and the reviewers for their constructive feedback.

\bibliographystyle{./bibliography/IEEEtran}
\bibliography{./bibliography/reinforce}

\begin{thebibliography}{10}
\providecommand{\url}[1]{#1}
\csname url@samestyle\endcsname
\providecommand{\newblock}{\relax}
\providecommand{\bibinfo}[2]{#2}
\providecommand{\BIBentrySTDinterwordspacing}{\spaceskip=0pt\relax}
\providecommand{\BIBentryALTinterwordstretchfactor}{4}
\providecommand{\BIBentryALTinterwordspacing}{\spaceskip=\fontdimen2\font plus
\BIBentryALTinterwordstretchfactor\fontdimen3\font minus
  \fontdimen4\font\relax}
\providecommand{\BIBforeignlanguage}[2]{{%
\expandafter\ifx\csname l@#1\endcsname\relax
\typeout{** WARNING: IEEEtran.bst: No hyphenation pattern has been}%
\typeout{** loaded for the language `#1'. Using the pattern for}%
\typeout{** the default language instead.}%
\else
\language=\csname l@#1\endcsname
\fi
#2}}
\providecommand{\BIBdecl}{\relax}
\BIBdecl

\bibitem{WOLFGANG20091642}
O.~Wolfgang, A.~Haugstad, B.~Mo, A.~Gjelsvik, I.~Wangensteen, and G.~Doorman,
  ``Hydro reservoir handling in norway before and after deregulation,''
  \emph{Energy}, vol.~34, no.~10, pp. 1642 -- 1651, 2009, 11th Conference on
  Process Integration, Modelling and Optimisation for Energy Saving and
  Pollution Reduction.

\bibitem{Helseth2016}
A.~{Helseth}, M.~{Fodstad}, and B.~{Mo}, ``Optimal medium-term hydropower
  scheduling considering energy and reserve capacity markets,'' \emph{IEEE
  Transactions on Sustainable Energy}, vol.~7, no.~3, pp. 934--942, 2016.

\bibitem{Sutton}
\BIBentryALTinterwordspacing
R.~S. Sutton and A.~G. Barto, \emph{{Reinforcement learning : an
  introduction}}.\hskip 1em plus 0.5em minus 0.4em\relax MIT Press, Cambridge,
  MA, 2018. [Online]. Available:
  \url{http://incompleteideas.net/book/the-book.html}
\BIBentrySTDinterwordspacing

\bibitem{Mnih2015}
V.~Mnih, K.~Kavukcuoglu, D.~Silver, A.~A. Rusu, J.~Veness, M.~G. Bellemare, and
  et~al., ``Human-level control through deep reinforcement learning,''
  \emph{Nature}, vol. 518, pp. 529 EP --, 02 2015.

\bibitem{Silver2016}
D.~Silver, A.~Huang, C.~J. Maddison, A.~Guez, L.~Sifre, G.~van~den Driessche,
  and et~al., ``Mastering the game of go with deep neural networks and tree
  search,'' \emph{Nature}, vol. 529, pp. 484 EP --, 01 2016.

\bibitem{Lillicrap2015}
\BIBentryALTinterwordspacing
T.~P. Lillicrap, J.~J. Hunt, A.~Pritzel, N.~Heess, T.~Erez, Y.~Tassa, and
  et~al., ``Continuous control with deep reinforcement learning,''
  \emph{Unpublished}, 2015. [Online]. Available:
  \url{https://arxiv.org/abs/1509.02971}
\BIBentrySTDinterwordspacing

\bibitem{rlblogpost}
A.~Irpan, ``Deep reinforcement learning doesn't work yet,''
  \url{https://www.alexirpan.com/2018/02/14/rl-hard.html}, 2018.

\bibitem{Watkins1989}
C.~J. C.~H. Watkins, ``{Learning from delayed rewards},'' PhD Thesis,
  University of Cambridge, 1989.

\bibitem{Watkins1992}
C.~J. C.~H. Watkins and P.~Dayan, ``{Q-learning},'' \emph{Machine Learning},
  vol.~8, no. 3-4, pp. 279--292, may 1992.

\bibitem{Haarnoja2018}
\BIBentryALTinterwordspacing
T.~Haarnoja, A.~Zhou, P.~Abbeel, and S.~Levine, ``Soft actor-critic: Off-policy
  maximum entropy deep reinforcement learning with a stochastic actor,''
  \emph{Unpublished}, 2018. [Online]. Available:
  \url{https://arxiv.org/abs/1801.01290}
\BIBentrySTDinterwordspacing

\bibitem{Haarnoja2018b}
\BIBentryALTinterwordspacing
T.~Haarnoja, A.~Zhou, K.~Hartikainen, G.~Tucker, S.~Ha, J.~Tan, and et~al.,
  ``Soft actor-critic algorithms and applications,'' \emph{Unpublished}, 2018.
  [Online]. Available: \url{https://arxiv.org/abs/1812.05905}
\BIBentrySTDinterwordspacing

\bibitem{Polyak1992}
B.~Polyak and A.~Juditsky, ``Acceleration of stochastic approximation by
  averaging,'' \emph{SIAM Journal on Control and Optimization}, vol.~30, no.~4,
  pp. 838--855, 1992.

\bibitem{Kumar2018}
\BIBentryALTinterwordspacing
V.~V. Kumar, ``{Soft actor critic demystified},'' Jan. 2018. [Online].
  Available: \url{https://github.com/vaishak2future/sac}
\BIBentrySTDinterwordspacing

\bibitem{NEURIPS2019_9015}
A.~Paszke, S.~Gross, F.~Massa, A.~Lerer, J.~Bradbury, G.~Chanan, and et~al.,
  ``Pytorch: An imperative style, high-performance deep learning library,'' in
  \emph{Advances in Neural Information Processing Systems 32}, H.~Wallach,
  H.~Larochelle, A.~Beygelzimer, F.~d\textquotesingle Alch\'{e}-Buc, E.~Fox,
  and R.~Garnett, Eds.\hskip 1em plus 0.5em minus 0.4em\relax Curran
  Associates, Inc., 2019, pp. 8024--8035.

\bibitem{Hahnloser2000}
R.~H.~R. Hahnloser, R.~Sarpeshkar, M.~A. Mahowald, R.~J. Douglas, and H.~S.
  Seung, ``Digital selection and analogue amplification coexist in a
  cortex-inspired silicon circuit,'' \emph{Nature}, vol. 405, no. 6789, pp.
  947--951, 2000.

\bibitem{Tieleman2012}
\BIBentryALTinterwordspacing
T.~Tieleman and G.~Hinton, ``Lecture 6.5 - rmsprop, coursera: Neural networks
  for machine learning,'' 2012. [Online]. Available:
  \url{https://www.cs.toronto.edu/~tijmen/csc321/slides/lecture_slides_lec6.pdf}
\BIBentrySTDinterwordspacing

\bibitem{Kingma2014}
\BIBentryALTinterwordspacing
D.~P. Kingma and J.~Ba, ``Adam: A method for stochastic optimization,''
  \emph{Unpublished}, 2014. [Online]. Available:
  \url{https://arxiv.org/abs/1412.6980}
\BIBentrySTDinterwordspacing

\bibitem{spinningup}
J.~Achiam, ``Spinning up in deep rl,''
  \url{https://spinningup.openai.com/en/latest/algorithms/sac.html}, 2018.

\bibitem{Cote2015}
P.~Côté and R.~Leconte, ``Comparison of stochastic optimization algorithms
  for hydropower reservoir operation with ensemble streamflow prediction,''
  \emph{Journal of Water Resources Planning and Management}, vol. 142, no.~2,
  p. 04015046, 2016.

\bibitem{Castelletti2009}
A.~Castelletti, S.~Galelli, M.~Restelli, and R.~Soncini-Sessa, ``Tree-based
  reinforcement learning for optimal water reservoir operation,'' \emph{Water
  Resources Research}, vol.~46, no.~9, 2010.

\bibitem{Lee2007}
J.-H. Lee and J.~W. Labadie, ``Stochastic optimization of multireservoir
  systems via reinforcement learning,'' \emph{Water Resources Research},
  vol.~43, no.~11, 2007.

\bibitem{Pan2010}
S.~J. {Pan} and Q.~{Yang}, ``A survey on transfer learning,'' \emph{IEEE
  Transactions on Knowledge and Data Engineering}, vol.~22, no.~10, pp.
  1345--1359, 2010.

\end{thebibliography}

\end{document}